\pdfoutput=1

\documentclass[11pt]{article}

\usepackage[]{ACL2023}

\usepackage{graphicx}
\usepackage{times}
\usepackage{latexsym}
\usepackage{amssymb}%
\usepackage{pifont}%
\newcommand{\cmark}{\ding{51}}%
\newcommand{\xmark}{\ding{55}}%
\usepackage{multirow}

\usepackage[T1]{fontenc}

\usepackage[utf8]{inputenc}

\usepackage{microtype}

\usepackage{inconsolata}
\usepackage{tabularx}
\usepackage{booktabs}
\usepackage{algorithm}
\usepackage{algpseudocode}

\newcommand{\eat}[1]{}

\title{Small Models are Valuable Plug-ins for Large Language Models}

\author{Canwen Xu$^1$\thanks{\ \ Work done during internship at Microsoft.}\ , Yichong Xu$^{2}$, Shuohang Wang$^{2}$, Yang Liu$^{2}$, \\ \textbf{Chenguang Zhu$^{2}$, Julian McAuley$^1$} \\
$^1$University of California, San Diego, $^2$Microsoft \\
$^1$\texttt{\{cxu,jmcauley\}@ucsd.edu}, $^2$\texttt{\{yicxu, shuowa, yaliu10, chezhu\}@microsoft.com}\\
}

\begin{document}
\maketitle
\begin{abstract}

Large language models (LLMs) such as GPT-3 and GPT-4 are powerful but their weights are often publicly unavailable and their immense sizes make the models difficult %
to be tuned with common hardware. As a result, effectively tuning these models with large-scale supervised data can be challenging. As an alternative, In-Context Learning (ICL) can only use a small number of supervised examples due to context length limits. In this paper, we propose Super In-Context Learning (SuperICL) which allows black-box LLMs to work with locally fine-tuned smaller models, resulting in superior performance on supervised tasks. Our experiments demonstrate that SuperICL can improve performance beyond state-of-the-art fine-tuned models while addressing the instability problem of in-context learning. Furthermore, SuperICL can enhance the capabilities of smaller models, such as multilinguality and interpretability.\footnote{Code available at \url{https://aka.ms/SuperICL}.}

\end{abstract}

\section{Introduction}
Large-scale pre-trained language models, such as GPT-3~\cite{gpt-3} and GPT-4~\cite{openai2023gpt4}, have demonstrated remarkable capabilities in a wide range of NLP tasks.
Despite the impressive performance of these recently released models, their size and limited accessibility of model weights can lead to difficulties in fine-tuning these models with supervised data, which is an effective way to adapt the models to specific tasks~\citep{liu2019roberta}.

An alternative approach, In-Context Learning (ICL, \citealp{gpt-3}), involves concatenating a few labeled examples with the test input, enabling the model to learn from the context. However, ICL is limited by the maximum context length of the LLM, restricting the number of examples it can utilize. Consequently, while ICL can usually perform few-shot learning with 16 or 32 examples, it cannot fully exploit supervised data when there are hundreds or thousands of examples.

To address these limitations, we propose Super In-Context Learning (SuperICL), a novel approach that enables black-box language models (e.g., GPT-3.5) to work with locally fine-tuned smaller models (e.g., RoBERTa, \citealp{liu2019roberta}), resulting in improved performance on supervised tasks. SuperICL is designed to overcome the challenges of poor performance and instability of ICL. %

\begin{figure*}
    \centering
    \includegraphics[width=0.85\linewidth]{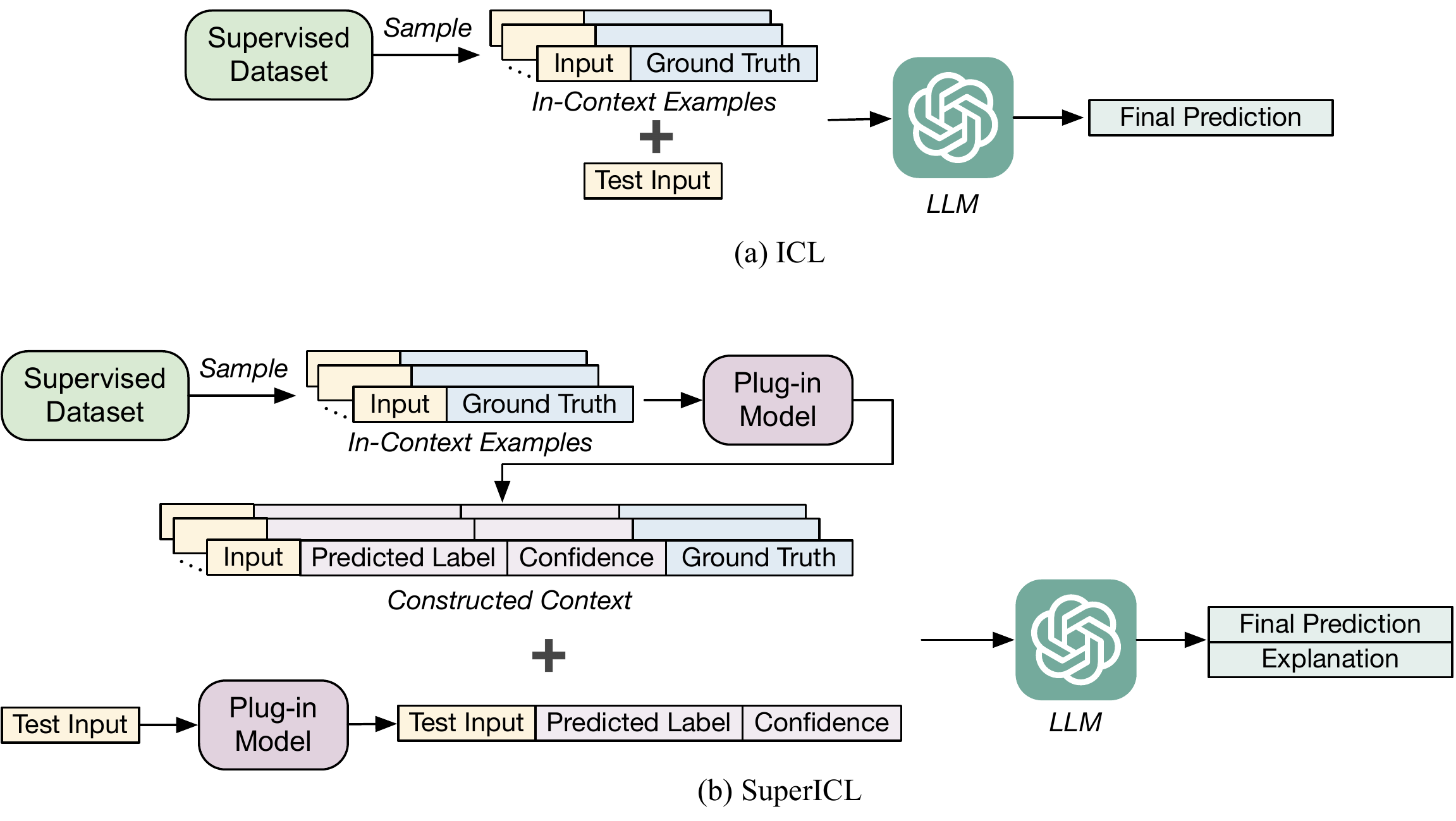}
    \caption{The workflow of ICL and SuperICL. There are three steps in SuperICL: (1) A context is constructed by randomly sampling from the training data and incorporating the plug-in model's predictions, including predicted labels and their corresponding confidence scores. (2) The test input is concatenated after the context, with the plug-in model's prediction attached. (3) Finally, a language model generates the final prediction along with an optional explanation.} 
    \label{fig:workflow}
\end{figure*}

SuperICL builds on the strengths of ICL while mitigating its limitations. As shown in Figure~\ref{fig:workflow}, SuperICL leverages a combination of an LLM with smaller models, which act as plug-ins, to perform supervised tasks efficiently. Specifically, we use the plug-in model to predict labels with confidence for in-context examples and concatenate them with the input text and ground-truth labels as context. For test examples, we also add the plug-in model's prediction and confidence to the test input and let the LLM  predict the final label and an explanation.
As these plug-in models have been fine-tuned on the task-specific data, they serve as a bridge between the large pre-trained model and the task-specific data, allowing for effective knowledge transfer and improved performance. %

We conduct extensive experiments to evaluate the effectiveness of SuperICL on GLUE~\citep{glue}, a standard benchmark for natural language understanding. Our results show that SuperICL: (1) achieves superior performance compared to state-of-the-art fine-tuned models and LLMs; (2) addresses the instability problem of ICL by allowing the plug-in models to absorb task-specific information while leaving the LLMs to focus on more general language understanding; (3) enhances the capabilities of plug-in models such as extending their multilinguality to cover a wider range of languages; (4) provides interpretability via the LLM by providing explanations for why it overrides predictions made by plug-in models. 

We then conduct a thorough analysis of how each component contributes to the final performance of SuperICL, as well as the impact from the number of in-context examples. We also explore the effects of adversarial attacks on plug-in models and how this affects SuperICL's performance. Our findings demonstrate the potential of combining large and small, cloud and local models, shedding light on a promising new paradigm for supervised learning in the era of large language models.

\section{Related Work}
\paragraph{In-Context Learning} Originally proposed in the GPT-3 paper~\citep{gpt-3}, In-Context Learning (ICL) is considered as a new paradigm that exploits LLMs on new tasks without updating the parameters of the model. It prepends few-shot training examples before the test input as a prompt, to enable large language models to find patterns and ``learn'' to predict. There have been successful applications of ICL in downstream tasks, such as Machine Translation~\citep{lin2021few, agrawal2022context} and data generation~\citep{progen}. 

Despite its success in few-shot learning, a major drawback of ICL is instability. The performance of ICL is sensitive to the selected in-context examples~\citep{zhao2021calibrate} and even their order~\citep{lu2022fantastically}. Based on these discoveries, there is a line of studies focused on constructing the context. LM-BFF~\citep{lmbff} and KATE~\citep{liu2022what} select training examples that are semantically similar to the test example. Another line of works~\citep{su2022selective, levy2022diverse, ye2023compositional} focus on mining diverse and representative examples from a training set. \citet{zhang2022active} utilizes active learning and reinforcement learning to select examples for ICL. Self-adaptive ICL~\citep{wu2022self,wu2023openicl} proposes a two-stage search framework to obtain the optimal in-context examples for each test input without using a separate validation set. Different from these works, SuperICL demonstrates that smaller models can be integrated into large language models for supervised tasks. Although it is orthogonal to these prior works, by fine-tuning the plug-in model with the entire training set, SuperICL reduces the necessity for selecting the optimal examples from the training set.

\begin{algorithm*}
\caption{Super In-Context Learning (SuperICL)}
\label{alg:supericl}
\begin{algorithmic}[1]
\Require Training set $D = \{(x_1, y_1), ..., (x_n, y_n)\}$, LLM $M$, a small pre-trained language model $P$
\Ensure{Predicted label $y_t$ and optional explanation $e_t$}

\State Fine-tune $P$ on $D$ to obtain the fine-tuned plug-in model $P'$
\State Randomly sample a set of examples $(x_i, y_i)$ from $D$ to be the set of in-context examples $D'$

\For{each example $(x_i, y_i)$ in $D'$}
\State Predict $y'_i$ and $c_i$ with $P'$ where $y'_i$ is the predicted label and $c_i$ is the confidence score
\EndFor
\State Construct the context $C$ by concatenating all $(x_i, y'_i, c_i, y_i)$

\For{each test example $x_t$}
\State Predict $y'_t$ and $c_t$ with $P'$ for the test example
\State Formulate complete input $I = C \oplus (x_t, y'_t, c_t)$ where $\oplus$ denotes concatenation
\State Use $M$ to predict $y_t$ from $I$
\State \textit{(Optional)} If $y_t \neq y'_t$, ask $M$ to generate an explanation $e_t$ for overriding the prediction of $P'$
\EndFor

\end{algorithmic}
\end{algorithm*}

Moreover, prior studies also investigate how to prepare language models for ICL. \citet{zhao2021calibrate} propose calibration with an empty test input to reduce the influence of the label distribution and ordering. MetaICL~\citep{min2022metaicl} meta-trains the language model to generalize to unseen tasks for better ICL performance. \citet{chen2022improving} propose four self-supervised objectives as intermediate tasks, to improve performance of language models on ICL. Notably, both \citet{min2022metaicl} and \citet{chen2022improving} require updating the weights, thus are not applicable to larger black-box models like GPT-3/4.

Besides studies aiming to improve ICL's performance, some studies have analyzed the underlying mechanism of ICL. \citet{min2022rethink} find that the label space, the distribution of the input text, and the overall format of the sequence are the key factors to ICL's performance.  They also claim that the ground labels are not significant to the performance of ICL, but this conclusion is contradicted by a later study~\citep{yoo2022ground}. Additionally, prior studies suggest ICL could be implicitly performing Bayesian inference~\citep{formulation_bayesian} or gradient descent~\citep{formulation_linear,formulation_gd_1,formulation_gd_2}.

\paragraph{Language Model Plug-ins} Large language models can exploit external tools to improve their capabilities. Toolformer~\citep{schick2023toolformer} introduces special symbols that allow the large language models to call external APIs to complete tasks. Visual ChatGPT~\citep{wu2023visual} plugs vision models into ChatGPT, allowing for multimodal generation. HuggingGPT~\citep{shen2023hugginggpt} uses ChatGPT to conduct task planning and select models according to their function descriptions available in Hugging Face, execute each subtask with the selected AI model, and summarize the response according to the execution results. Different from these works, our work is under a classic supervised learning and demonstrates that even tasks like text classification, which is sometimes considered ``solved'' by smaller language models, can still benefit from combination with a large language model.

\begin{table*}[t]
\small
\centering
\begin{tabularx}{\linewidth}{lX}
\toprule
\textbf{\textit{(a) Context}} &
  \texttt{Sentence 1:} Federal agent Bill Polychronopoulos said it was not known if the man, 30, would be charged. \newline
\texttt{Sentence 2:} Federal Agent Bill Polychronopoulos said last night the man involved in the Melbourne incident had been unarmed. \newline
\texttt{RoBERTa-Large Prediction:} equivalent (Confidence: 0.51) \newline
\texttt{Label:} not\_equivalent \newline

\texttt{Sentence 1:} Five more human cases of West Nile virus, were reported by the Mesa County Health Department on Wednesday. \newline
\texttt{Sentence 2:} As of this week, 103 human West Nile cases in 45 counties had been reported to the health department. \newline
\texttt{RoBERTa-Large Prediction:} not\_equivalent (Confidence: 0.98) \newline
\texttt{Label:} not\_equivalent \newline
...
\\
\midrule
\textbf{\textit{(b) Test Input}} & 
\texttt{Sentence 1:} Cooley said he expects Muhammad will similarly be called as a witness at a pretrial hearing for Malvo. \newline
\texttt{Sentence 2:} Lee Boyd Malvo will be called as a witness Wednesday in a pretrial hearing for fellow sniper suspect John Allen Muhammad. \newline
\texttt{RoBERTa-Large Prediction:} equivalent (Confidence: 0.82) \\
\midrule
\textbf{\textit{(c) Label Prediction}} & \texttt{Label:} not\_equivalent \\
\midrule
\textbf{\textit{(d) Explanation}} & \texttt{Explanation for overriding the prediction:} The two sentences are talking about different people, John Allen Muhammad and Lee Boyd Malvo, and thus the prediction should be not\_equivalent. 
\\
\bottomrule  
\end{tabularx}
\caption{An example of the constructed context and inference procedure from the MRPC dataset. We first construct the context by sampling from the supervised dataset and attach the plug-in model's predictions. Then, for each test example, we ask the large language model to predict the label based on the input and the plug-in model's prediction. We use a prompt to ask the model to explain the decision if the label predicted by the plug-in model is overridden. The text field names (e.g., Sentence 1) are the original field names provided in the dataset. 
\label{tab:example}}
\end{table*}

\section{Super In-Context Learning}
Super In-Context Learning (SuperICL) combines LLMs with locally fine-tuned smaller models, allowing them to work together to improve performance on supervised tasks. The smaller models act as plug-ins, providing task-specific knowledge and predictions, while the large pre-trained models focus on general language understanding. The overall workflow of SuperICL is shown in Figure~\ref{fig:workflow} and the complete algorithm is depicted in Algorithm \ref{alg:supericl}.

\paragraph{Plug-in Model Fine-tuning} The first step in the SuperICL process is fine-tuning a small NLP model, e.g., RoBERTa~\citep{liu2019roberta}, on task-specific labeled data. 
This fine-tuning process on the entire training data is made possible due to the smaller size of the model and its local accessibility.
This is in contrast to ICL, whose usage of labeled data is severely limited by the LLM's context length.
The fine-tuned small model is then integrated as a plug-in for the LLM in subsequent steps as follows. 

\paragraph{Context Construction} Next, a context is constructed for the LLM to utilize the task-specific knowledge provided by the smaller model. This context consists of a set of examples randomly sampled from the training data, along with their corresponding predictions by the smaller plug-in model. The predictions include both the predicted labels and their associated confidence scores. An example is shown in Table~\ref{tab:example}.

On one hand, by incorporating the predicted labels by the plug-in model, the LLM can better understand the relationship among the input examples, ground-truth labels and the plug-in model's expertise. This will help the LLM in the subsequent decision-making process to produce final predictions.
On the other hand, confidence scores provide a measure of the plug-in model's uncertainty in its predictions. By incorporating these scores in the context, the LLM can trust predictions where the plug-in model is highly confident and be more cautious when the plug-in model is uncertain. Furthermore, confidence scores can help guide the LLM's attention towards in-context examples that are more challenging, enabling it to learn from these difficult cases and potentially improve its overall performance on the task. %

In summary, by considering both the predicted label and the associated confidence from the plug-in model, the LLM decides whether to follow the given predictions or to rely on its own understanding of the task, leading to more accurate predictions overall.

\paragraph{Inference} Once the context has been constructed, the test input (an example shown in Table~\ref{tab:example}(b)) is concatenated with the context, forming a complete input for the large language model. The plug-in model's prediction for the test input, including the predicted label and confidence score, is also attached to the input. 
Thus, the LLM's input includes the context, test input, and plug-in model's prediction. The LLM then generates a final prediction for the test input, as shown in Table~\ref{tab:example}(c). Optionally, as shown in Table~\ref{tab:example}(d), the LLM can also provide an explanation for its prediction, giving insight into why it chose to override or follow the plug-in model's prediction. This additional interpretability can be valuable for understanding the decision-making process of the combined SuperICL model.

\section{Experiments}

\begin{table*}[t]
\resizebox{\linewidth}{!}{
\begin{tabular}{lccccccccc}
\toprule
Methods            & MNLI-m & MNLI-mm & SST-2 & QNLI  & MRPC  & QQP & CoLA & RTE   & Avg.      \\
\midrule
GPT-3.5 ICL           & 80.80   & 82.39   & 91.39 & 80.52 & 60.05 & 81.64  & 60.51       & 86.28 & 81.32 \\
RoBERTa-Large & 88.68  & 89.47   & 96.44 & 94.07 & 83.09 & 92.11  & 64.55       & 87.00    & 88.68 \\
\midrule
SuperICL       & \textbf{89.31}  & \textbf{89.61}   & \textbf{96.79} & \textbf{94.16} & \textbf{86.03} & \textbf{92.14}  & \textbf{64.57}       & \textbf{87.73} & \textbf{89.90} \\
\bottomrule
\end{tabular}
}
\caption{\label{tab:glue} Experimental results on GLUE~\citep{glue} development set. The metric for CoLA is Matthews Correlation and all other tasks use accuracy.} 
\end{table*}

\begin{table}[t]
\small
\centering
\begin{tabular}{cccc}
\toprule
Lang.     & GPT-3.5 ICL & XLM-V & SuperICL \\
\midrule
en   & 74.03       & 83.55 & \textbf{83.87}    \\
ar   & 60.15       & 70.78 & \textbf{72.28}    \\
bg   & 67.64       & 77.09 & \textbf{77.74}    \\
de   & 71.78       & 75.23 & \textbf{80.28}    \\
el   & 65.85       & 72.73 & \textbf{74.29}    \\
es   & 76.79       & 77.07 & \textbf{81.38}    \\
fr   & 74.99       & 77.01 & \textbf{77.47}    \\
hi   & 56.29       & 69.62 & \textbf{70.02}    \\
ru   & 65.39       & 73.53 & \textbf{76.85}    \\
sw   & 56.13       & 67.43 & \textbf{68.94}    \\
th   & 57.03       & 68.90 & \textbf{69.36}    \\
tr   & 66.01       & 72.34 & \textbf{72.63}    \\
ur   & 51.18       & \textbf{63.57} & 57.90     \\
vi   & 62.91       & 72.91 & \textbf{74.45}    \\
zh   & 67.90       & 73.75 & \textbf{74.21}    \\
\midrule
Avg. & 64.94       & 73.03 & \textbf{74.11}   \\
\bottomrule
\end{tabular}
\caption{\label{tab:xnli} Experimental results on the XNLI~\cite{conneau2018xnli} test set. The metric is accuracy.}
\end{table}

\subsection{Experimental Settings}
\paragraph{Benchmarks} We focus on the full supervised setting, where we have access to the entire training set. We conduct experiments on two widely-used benchmarks: the GLUE benchmark~\citep{glue} for natural language understanding tasks and the XNLI benchmark~\citep{conneau2018xnli} for zero-shot cross-lingual natural language inference tasks, where the models are trained on English and tested on other languages. Our goal is to examine the learning ability of SuperICL on standard benchmarks and whether it can empower smaller models with its multilingual capability. For both ICL and SuperICL, we only consider the prediction to be correct when the generated label matches the predefined label exactly. For analytical experiments, we evaluate the model on a subset of GLUE consisting of three representative tasks: MNLI, SST-2 and MRPC, due to budget constraints.

\paragraph{Large Language Model and Plug-ins} We use OpenAI's \texttt{text-davinci-003} language model, also known as GPT-3.5. For the GLUE benchmark, we use RoBERTa-large~\citep{liu2019roberta} as the plug-in model. For the XNLI benchmark, we use XLM-V~\citep{liang2023xlm}, as the plug-in model. Both models are fine-tuned on their respective tasks to serve as plug-ins for SuperICL. For GLUE tasks, we randomly select 32 examples from the training set. For XNLI, as the input is multilingual, the BPE tokenizer used in GPT-3.5 results in a longer token sequence. Thus, we use at most 16 examples for each language. Note that for some languages (e.g., Thai), the in-context examples are fewer than 16, as we fit as many examples as possible to the maximum allowed sequence length of 4,096 of GPT-3.5. For our main experiments and all analysis experiments, we compare the performance of SuperICL with ICL~\citep{gpt-3} on the same selection of in-context examples and the predictions made by the plug-in models alone. 

\subsection{Main Results}
\paragraph{GLUE} As shown in Table~\ref{tab:glue}, SuperICL outperforms both GPT-3.5 ICL and the plug-in model RoBERTa-Large with an average advantage of 8.58 and 1.22 on GLUE, respectively. It is worth noting that SuperICL consistently outperforms the baselines on all tasks, which makes it a reliable choice that would not compromise the performance of the plug-in model. 

\paragraph{XNLI} 
For XNLI, as presented in Table~\ref{tab:xnli}, while XLM-V~\citep{liang2023xlm} is specifically designed for multilingual tasks, combining it with GPT-3.5 can still lead to significant improvements in most languages. However, SuperICL fails to enhance the performance of XLM-V for Urdu. It is worth mentioning that GPT-3.5 ICL also exhibits poor performance for Urdu, implying that GPT-3.5 may lack ability for low-resource languages like Urdu. This is also consistent with recent analysis on the multilinguality of GPT-3.5/ChatGPT~\citep{lai2023chatgpt}. Additionally, since the BPE tokenizer used in GPT-3.5 yields more tokens for non-Latin languages, the number of in-context examples is limited, adversely affecting the model's performance. We believe that subsequent GPT models that employ a multilingual tokenizer, train on more non-English data, and have a longer maximum context can achieve even better performance for cross-lingual SuperICL.

\subsection{Ablation Study}

\begin{table}[t]
\centering
\resizebox{\linewidth}{!}{
\begin{tabular}{cllllccc}
\toprule
& \multicolumn{3}{c}{Components}                        & \multirow{2}{*}{MNLI} & \multirow{2}{*}{SST-2} & \multirow{2}{*}{MRPC}  \\
\cmidrule{2-4}
& \textit{(a) Ctxt.} & \textit{(b) Conf.} & \textit{(c) Ref.}  \\
\midrule
\multicolumn{4}{l}{GPT-3.5 ICL}         & 80.80  & 91.39 & 60.05 \\
\multicolumn{4}{l}{RoBERTa-Large}       & 88.68  & 96.44 & 83.09 \\

\midrule
(1) & \cmark & \cmark & \xmark         & 81.23  & 92.43 & 65.69 \\
(2) & \cmark & \xmark & \cmark         & 88.75  & 96.67 & 83.09 \\
(3) & \xmark & \cmark & \cmark         & 88.89  & 96.44 & 83.59 \\
(4) & \xmark & \xmark & \cmark         & 88.84  & 96.44 & 83.58 \\
\midrule
 & \cmark & \cmark & \cmark         & \textbf{89.31}  & \textbf{96.79} & \textbf{86.03} \\
\bottomrule
\end{tabular}
}
\caption{\label{tab:ablation}Experimental results of the ablation study. \textit{(a) Ctxt.} means the in-context examples from the training set; \textit{(b) Conf.} represents the plug-in model's confidence score; \textit{(c) Ref.} means whether we use the plug-in model's prediction for the test input.}
\end{table}

We conduct an ablation study to understand the effect of each component in SuperICL. We investigate the performance of the three components in SuperICL: (a) Context, which comprises in-context examples; (b) The confidence scores of the plug-in model in both in-context examples and test input; (c) The plug-in model's prediction for the test input.

The experimental results are shown in Table~\ref{tab:ablation}: (1) We first attempt to remove the plug-in model's prediction for the test input. This has a significant negative impact on the performance of ICL as it creates a mismatch between in-context examples and test input. Interestingly, even though we remove the plug-in model for test input, SuperICL can still outperform ICL. We suspect this is due to an in-context effect similar to knowledge distillation~\citep{hinton2015distilling}, which transfers task knowledge from the fine-tuned RoBERTa to GPT-3.5. (2) We attempt to remove the confidence scores from SuperICL which results in a decrease in its performance. This is because GPT-3.5 becomes unaware of the uncertainty of RoBERTa, and as a result, it is unable to determine when to override the prediction. Also, similar to removing the softmax score from knowledge distillation, removing the confidence score makes knowledge transfer less effective. (3) When removing all in-context examples, SuperICL is essentially doing zero-shot inference for the test input. Although there is a slight improvement over RoBERTa, we can see adding in-context examples help SuperICL learn to calibrate the confidence and override RoBERTa's prediction. Also similar to ICL versus zero-shot inference, adding in-context examples helps GPT-3.5 to improve its own task-specific performance. (4) Further removing confidence scores from zero-shot inference also slightly decreases the performance.

\subsection{Analysis on Prediction Overrides}

We also analyze the statistics of predictions overridden by GPT-3.5, as displayed in Table~\ref{tab:flip}. A significant difference can be observed between various datasets. On both MNLI and SST-2, GPT-3.5 overrides only a minimal portion of examples (approximately 0.2\%), but with high accuracy. Conversely, GPT-3.5 overrides a substantial percentage of 12.5\% of predictions made by RoBERTa, although with lower accuracy. These findings suggest that the override behavior of SuperICL is heavily reliant on the specific dataset and the performance of the plug-in model.

\begin{table}[t]
\resizebox{\linewidth}{!}{
\begin{tabular}{lrrr}
\toprule
Method      & MNLI & SST-2 & MRPC   \\
\midrule
\%Overriden           & 0.22\%    &  0.23\%   & 12.50\%   \\
Overridden Accuracy & 81.81\% &   100.00\%  & 64.71\%\\
\bottomrule
\end{tabular}
}
\caption{\label{tab:flip} Statistics of overridden predictions. ``\%Overridden'' indicates the percentage of final predictions that differ from the plug-in model's predictions, out of the total number of examples. ``Overridden Accuracy'' represents the percentage of correct predictions among the overridden ones. }
\end{table}

\begin{figure}[t]
    \centering
    \includegraphics[width=\columnwidth]{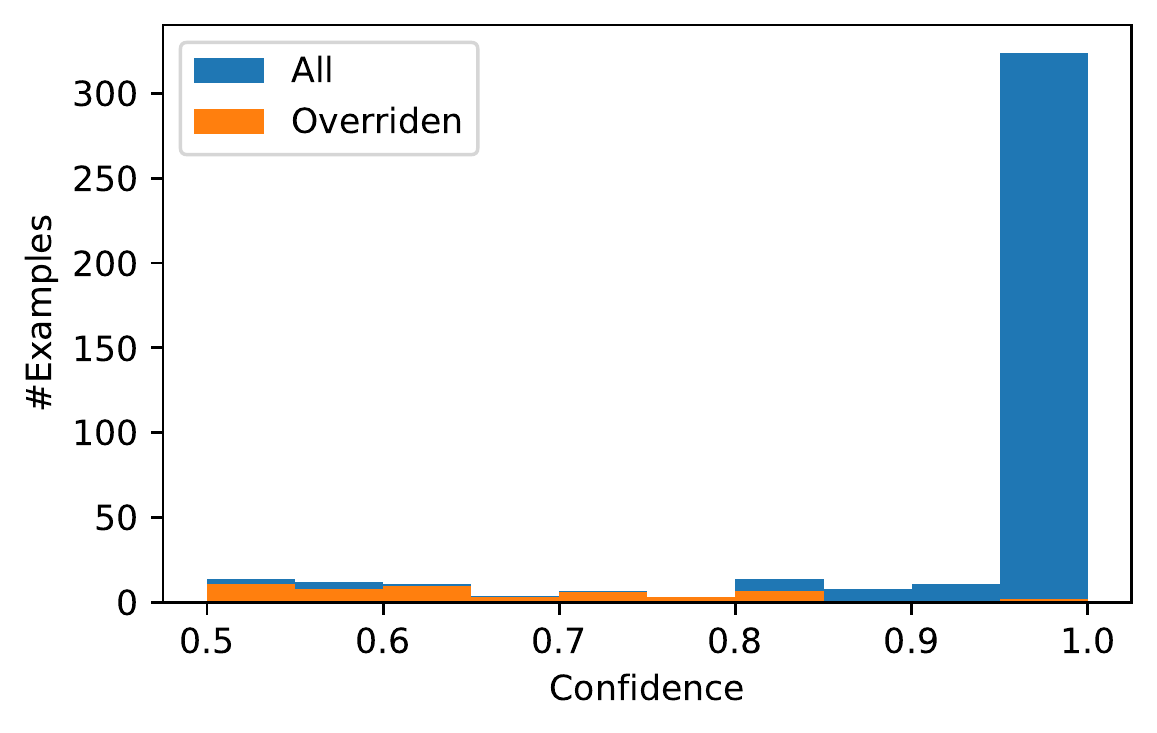}
    \caption{Effect of plug-in model confidence for overrides. The figure is the distribution of RoBERTa confidence for all examples (blue) and examples with a final prediction overridden by GPT-3.5 (orange) on MRPC. }
    \label{fig:conf_distrib}
\end{figure}

To gain insights into the decision-making process of the LLM in overriding the predictions of the plug-in model, we examine the distribution of confidence levels exhibited by RoBERTa and the extent to which GPT-3.5 overrides them, as shown in Figure~\ref{fig:conf_distrib}. The findings reveal a pattern where GPT-3.5 tends to override predictions when the plug-in model's confidence is low. This behavior supports our motivation and indicates that GPT-3.5 recognizes the uncertainty associated with the plug-in model's predictions via the confidence scores.

\begin{table}[t]
\resizebox{\linewidth}{!}{
\begin{tabular}{clcccccc}
\toprule
& \multirow{2}{*}{Method} &              \multicolumn{5}{c}{Random seed}                                                           & \multirow{2}{*}{Var.} \\
\cmidrule{3-7}
                         &         & 42             & 0              & 1              & 2              & 3              & \multicolumn{1}{c}{}                          \\
\midrule
\multirow{3}{*}{\rotatebox[origin=c]{90}{MNLI}}    & ICL           & 80.80           & 81.26          & 79.74          & 81.26          & 80.79          & 0.39                                          \\
                         & RoBERTa & 88.68          & 88.68          & 88.68          & 88.68          & 88.68          & -                                             \\
                         & SuperICL      & \textbf{89.31} & \textbf{88.94} & \textbf{88.79} & \textbf{89.17} & \textbf{88.78} & 0.06                                          \\
\midrule
\multirow{3}{*}{\rotatebox[origin=c]{90}{SST-2}}   & ICL           & 91.39          & 94.04          & 94.38          & 93.12          & 93.46          & 1.35                                          \\
                         & RoBERTa & 96.44          & 96.44          & 96.44          & 96.44          & 96.44          & -                                             \\
                         & SuperICL      & \textbf{96.79} & \textbf{96.56} & \textbf{96.56} & \textbf{96.56} & \textbf{96.56} & 0.01                                          \\
\midrule
\multirow{3}{*}{\rotatebox[origin=c]{90}{MRPC}}    & ICL           & 60.05          & 73.53          & 73.28          & 73.28          & 65.44          & 37.50                                         \\
                         & RoBERTa & 83.09          & 83.09          & 83.09          & 83.09          & 83.09          & -                                             \\
                         & SuperICL      & \textbf{86.03} & \textbf{87.99} & \textbf{87.75} & \textbf{84.31} & \textbf{86.52} & 2.20 \\
\bottomrule
\end{tabular}
}
\caption{\label{tab:selection}Accuracy and variance of ICL and SuperICL with example selections randomly sampled with different seeds.}
\end{table}

\begin{figure*}
\centering
    \includegraphics[width=0.32\linewidth]{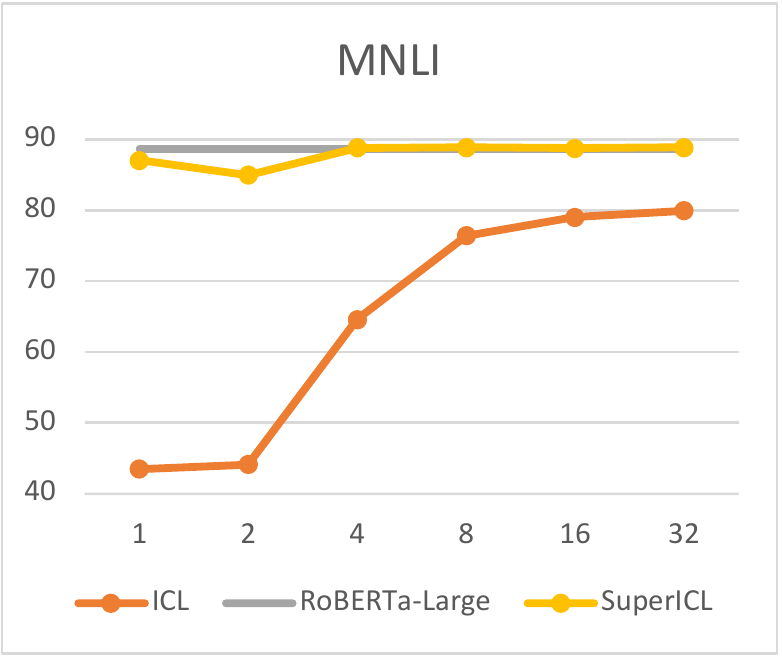}
    \hfill
    \includegraphics[width=0.32\linewidth]{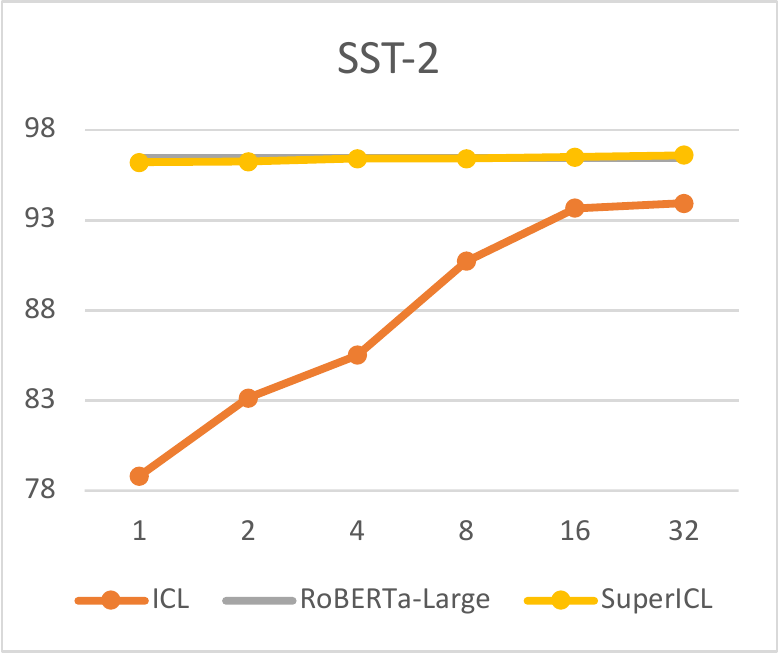}
    \hfill
    \includegraphics[width=0.32\linewidth]{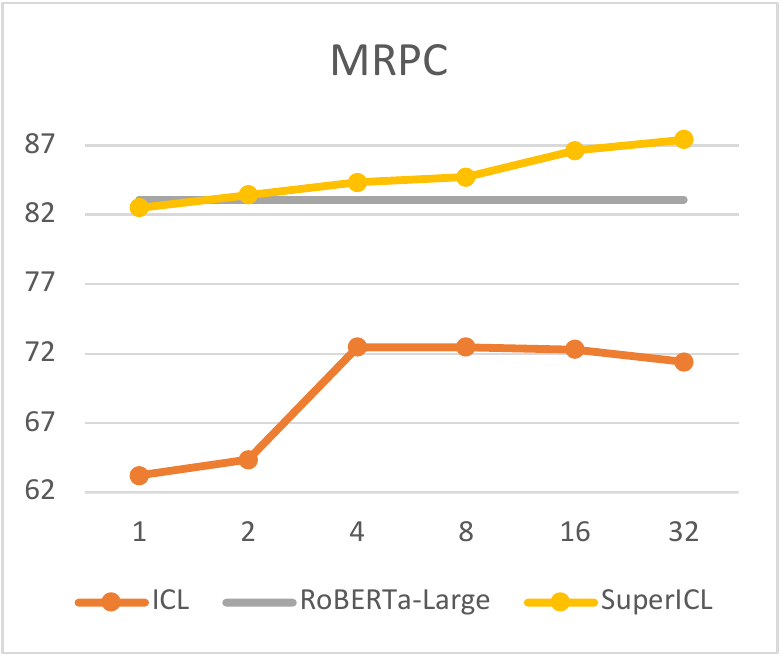}
\caption{\label{fig:num} Effect of number of examples on the performance of ICL and SuperICL. The results are averages of three runs.}
\end{figure*}

\subsection{Analysis on Example Selection}

We compare ICL and SuperICL by analyzing the sensitivity of different in-context examples. To ensure a fair comparison, we randomly sample five batches of in-context examples for each dataset using different random seeds, ensuring that the same in-context examples are used for both ICL and SuperICL. 

Our results, as shown in Table~\ref{tab:selection} show that ICL demonstrates a larger variance compared to SuperICL, especially on MRPC, and its performance is drastically affected by the selection of in-context examples. On the other hand, SuperICL consistently outperforms both ICL and RoBERTa, while maintaining a more stable performance.

We argue that SuperICL's improved stability is due to the added reference prediction from the plug-in model, which has been trained on a large supervised training set. This helps the LLM to focus on learning to correct RoBERTa's predictions and override them when needed, minimizing the differences caused by different in-context examples. 

\begin{table}[t]
\resizebox{\linewidth}{!}{
\begin{tabular}{lccc}
\toprule
Method      & MNLI & SST-2 & MRPC   \\
\midrule
ICL           & 80.80    & 91.39    & 60.05   \\
\midrule
RoBERTa-Large & 88.68    & 96.44    & 83.09   \\
SuperICL + RoBERTa & \textbf{89.31} & \textbf{96.79} & \textbf{86.03} \\
\midrule
DeBERTa V3-Large & 90.49    & 96.56    & 90.44   \\
SuperICL + DeBERTa      & \textbf{90.76}    & \textbf{96.79}    & \textbf{90.93}   \\
\bottomrule
\end{tabular}
}
\caption{\label{tab:plugin} Experimental results of SuperICL with different plug-in models.}
\end{table}

\begin{table}[t]
\resizebox{\linewidth}{!}{
\begin{tabular}{lcccc}
\toprule
Test Set      & R1 & R2 & R3 & All   \\
\# Examples    & 1000    & 1000    & 1200    & 3200  \\
\midrule
ICL           & 59.50    & 52.40    & 52.58   & 54.69 \\
RoBERTa-Large & 41.60    & 27.40    & 24.58   & 30.78 \\
SuperICL      & 56.10    & 42.70    & 44.17   & 47.44 \\
\bottomrule
\end{tabular}
}
\caption{\label{tab:anli} Zero-shot results on ANLI~\citep{nie2019adversarial}. ICL and SuperICL use in-context examples sampled from MNLI. The RoBERTa-Large model is fine-tuned on MNLI. R1, R2 and R3 denote the first, second and third round of adversarial attacks, respectively.}
\end{table}

\begin{table*}[t]
\small
\centering
\begin{tabularx}{\linewidth}{X}
\toprule
\textbf{\textit{(1) MNLI}} \\
\texttt{Premise}: Then he sobered. \newline
\texttt{Hypothesis}: He was always sober.
\newline
\texttt{RoBERTa Prediction}: neutral (Confidence: 0.7) \xmark
\newline
\texttt{Final Prediction}: contradiction \cmark \newline
\texttt{Explanation:} The premise states that he sobered, implying that he was not sober before, while the hypothesis states that he was always sober, which contradicts the premise. 
\\
\midrule
\textbf{\textit{(2) MNLI}} \\
\texttt{Premise}: Without the discount, nobody would buy the stock. \newline
\texttt{Hypothesis}: Nobody would buy the stock if there was a discount. \newline
\texttt{RoBERTa Prediction}: contradiction (Confidence: 0.6) \cmark \newline
\texttt{Final Prediction}:  entailment \xmark \newline
\texttt{Explanation:} The premise and hypothesis are essentially the same statement, but with the premise being a negative statement and the hypothesis being a positive statement. The model interprets this as a contradiction, but it should be an entailment. \\
\midrule
\textbf{\textit{(3) SST-2}} \\
\texttt{Sentence}: no screen fantasy-adventure in recent memory has the showmanship of clones' last 45 minutes. \newline 
\texttt{RoBERTa Prediction}: negative (Confidence: 0.54) \xmark \newline
\texttt{Final Prediction}: positive \cmark \newline
\texttt{Explanation:} The sentence is actually a positive statement, praising the last 45 minutes of the movie for its showmanship. The prediction was flipped because the sentence also contains the word "no" which can be interpreted as a negative sentiment. \\

\bottomrule  
\end{tabularx}
\caption{Examples of SuperICL's predictions and explanations. The contexts are omitted. The prompts here are modified for clarity. The original prompt template is shown in Table~\ref{tab:example}. The correct and incorrect predictions are marked with \cmark and \xmark, respectively. 
\label{tab:case}}
\end{table*}

\subsection{Analysis on Number of Examples}
We compare the effect of the number of examples on the performance of ICL and SuperICL in Figure~\ref{fig:num}. As shown, SuperICL can already achieve the best accuracy with merely 2 or 4 examples, while ICL requires more examples to achieve reasonable accuracy. Also, on the more difficult dataset MRPC, SuperICL can improve the performance of RoBERTa by a large margin with more in-context examples present. 

\subsection{Analysis on Plug-in Models}
We also attempt to replace the plug-in model with a state-of-the-art model, DeBERTa V3-Large~\citep{he2021debertav3} fine-tuned on the datasets. Our results, presented in Table~\ref{tab:plugin}, demonstrate that SuperICL continues to enhance the performance of state-of-the-art models, although the improvement is smaller compared to RoBERTa. This reduction may be attributed to the smaller capability gap between the small and large models. However, we expect that using an even more advanced large language model in the future will resolve this issue.

\subsection{Analysis on Adversarial Robustness}
\label{sec:adv}
Additionally, we analyze the adversarial robustness of SuperICL by testing it on ANLI~\citep{nie2019adversarial}. ANLI is a dataset for evaluating the robustness and generalization of natural language inference (NLI) models. It consists of 16,000 premise-hypothesis pairs that are categorized into three classes: entailment, contradiction, and neutral. The dataset is constructed with three rounds (R1, R2, and R3) and thus has three splits, with R3 being the most challenging and diverse. ANLI is collected using a human-and-model-in-the-loop training method, where human annotators act as adversaries and attempt to fool the model into misclassifying while still being understandable to other humans. This benchmark is designed to be challenging for language models including RoBERTa as RoBERTa is attacked in R2 and R3 of data construction.

As shown in Table~\ref{tab:anli}, GPT-3.5 ICL is rather robust while RoBERTa-Large is vulnerable to adversarial attack. However, this directly has a negative impact on SuperICL. Although SuperICL achieves better performance than RoBERTa-Large, it underperforms ICL. This finding suggests that SuperICL's performance relies on the performance of the incorporated plug-in model and adversarial attack to the plug-in model could lead to a drastic performance drop for SuperICL.

\section{Case Study}
We conduct a case study to better understand the behavior of SuperICL, with three examples presented in Table~\ref{tab:case}. We find that even without any explicit task instruction, GPT-3.5 demonstrates the ability to comprehend the tasks and explain its own reasoning. In the first example from Table~\ref{tab:case}, GPT-3.5 effectively grasps the implication in the premise that ``he'' was not sober. However, in the second example, GPT-3.5 incorrectly flips the prediction, possibly due to confusion caused by negation. This phenomenon has been recognized as a common flaw in LLMs, as noted by \citet{hosseini2021understanding} and \citet{jang2023can}. In the last example, GPT-3.5 not only corrects RoBERTa's prediction successfully but also provides an analysis explaining why RoBERTa makes the wrong prediction.

\section{Conclusion and Future Work}
In this paper, we propose SuperICL, a simple yet effective method for combining a large language model API with a locally fine-tuned plug-in model. For future work, we would like to explore using large language models to plan for the fine-tuning of the local plug-in model for an unseen task and automate the entire workflow. Also, a theoretical analysis may be important to further reveal the internal mechanism of SuperICL.

\section*{Limitations}
\paragraph{Additional Delay and Cost} Since SuperICL involves serialized small and large models, the total inference delay equals to the sum of the inference delay of the two models. Also, calling the API of large language model can be expensive compared to using a locally deployed small model.

\paragraph{Adversarial Vulnerability} As discussed in Section~\ref{sec:adv}, the vulnerability of the plug-in model to adversarial attacks can be inherited by SuperICL. Thus, when the plug-in model is under adversarial attack, the entire system could underperform ICL.

\paragraph{Limited Evaluation Tasks} Due to space and budget limit, we only investigate text classification in this paper. However, it would be interesting to also look into generation tasks such as text summarization, question answering, and semantic parsing. 

\section*{Broader Impact} 
As a technique that combines large and small language models for improved predictions, SuperICL shares the potential social biases of language models. While our approach is not likely to amplify these biases compared to other methods, it is important to investigate whether SuperICL has any effect on increasing or decreasing them. Furthermore, incorporating small models as plug-ins to the inference of large language models may lead to a slightly higher carbon footprint, resulting in a negative environmental impact. Therefore, practitioners should carefully consider the trade-offs between performance gains and environmental costs when using SuperICL.

\section*{Acknowledgements}
We would like to thank Junheng Hao, Ziyi Yang, Dan Iter and Daya Guo for discussion.

\bibliography{custom}
\bibliographystyle{acl_natbib}

\end{document}